\pdfoutput=1

\documentclass[11pt]{article}

\usepackage[table]{xcolor}
\usepackage{tcolorbox}
\tcbuselibrary{skins,breakable}
\usepackage{booktabs}
\usepackage{adjustbox}
\usepackage{array}
\usepackage{multirow}
\usepackage{CJKutf8}

\usepackage[preprint]{acl}

\usepackage{times}
\usepackage{latexsym}

\usepackage[T1]{fontenc}

\usepackage[utf8]{inputenc}

\usepackage{microtype}

\usepackage{inconsolata}

\usepackage{graphicx}

\title{From Black Box to Transparency: Enhancing Automated Interpreting Assessment with Explainable AI in College Classrooms}

\author{%
    Zhaokun Jiang\textsuperscript{1} \quad\
    Ziyin Zhang\textsuperscript{1}\thanks{daenerystargaryen@sjtu.edu.cn} \
    \vspace{6pt}\\
    \textsuperscript{1}Shanghai Jiao Tong University \\
}

\begin{document}
\maketitle
\begin{abstract}
Recent advancements in machine learning have spurred growing interests in automated interpreting quality assessment. Nevertheless, existing research suffers from insufficient examination of language use quality, unsatisfactory modeling effectiveness due to data scarcity and imbalance, and a lack of efforts to explain model predictions. To address these gaps, we propose a multi-dimensional modeling framework that integrates feature engineering, data augmentation, and explainable machine learning. This approach prioritizes explainability over ``black box'' predictions by utilizing only construct-relevant, transparent features and conducting Shapley Value (SHAP) analysis. Our results demonstrate strong predictive performance on a novel English-Chinese consecutive interpreting dataset, identifying BLEURT and CometKiwi scores to be the strongest predictive features for fidelity, pause-related features for fluency, and Chinese-specific phraseological diversity metrics for language use. Overall, by placing particular emphasis on explainability, we present a scalable, reliable, and transparent alternative to traditional human evaluation, facilitating the provision of detailed diagnostic feedback for learners and supporting self-regulated learning advantages not afforded by automated scores in isolation.
\end{abstract}

\section{Introduction}

Interpreting, or oral translation, is a complex yet pivotal linguistic competency that offers extensive educational benefits by fostering advanced linguistic, communicational, cognitive, and emotional capabilities~\citep{Pchhacker2001,Gile2021}. It enhances active listening~\citep{Lee2013}, oral proficiency~\citep{Han-Lu2025}, vocabulary acquisition~\citep{Chen2024}, and cross-cultural communication~\citep{Stachl-Peier2020}, while also strengthening higher-order cognitive functions~\citep{Dong-Xie2014} and anxiety management capabilities~\citep{Zhao2022}.

Given its multifaceted benefits, interpreting has increasingly been recognized as both a valuable pedagogical tool and the “fifth skill”~\citep{Mellinger2018} alongside listening, speaking, reading, and writing. The intricate nature of interpreting necessitates a continuous cycle of structured practice, rigorous assessment, and diagnostic feedback~\citep{Gile2021}. However, traditional human-based assessment often requires raters to simultaneously consult the source text, the interpreted output, and detailed rating scales, a cognitively demanding process that increases the risk of scoring bias and inconsistency~\citep{Lee2019,Han-etal2024}.

\begin{figure*}
    \centering
    \includegraphics[width=0.42\linewidth]{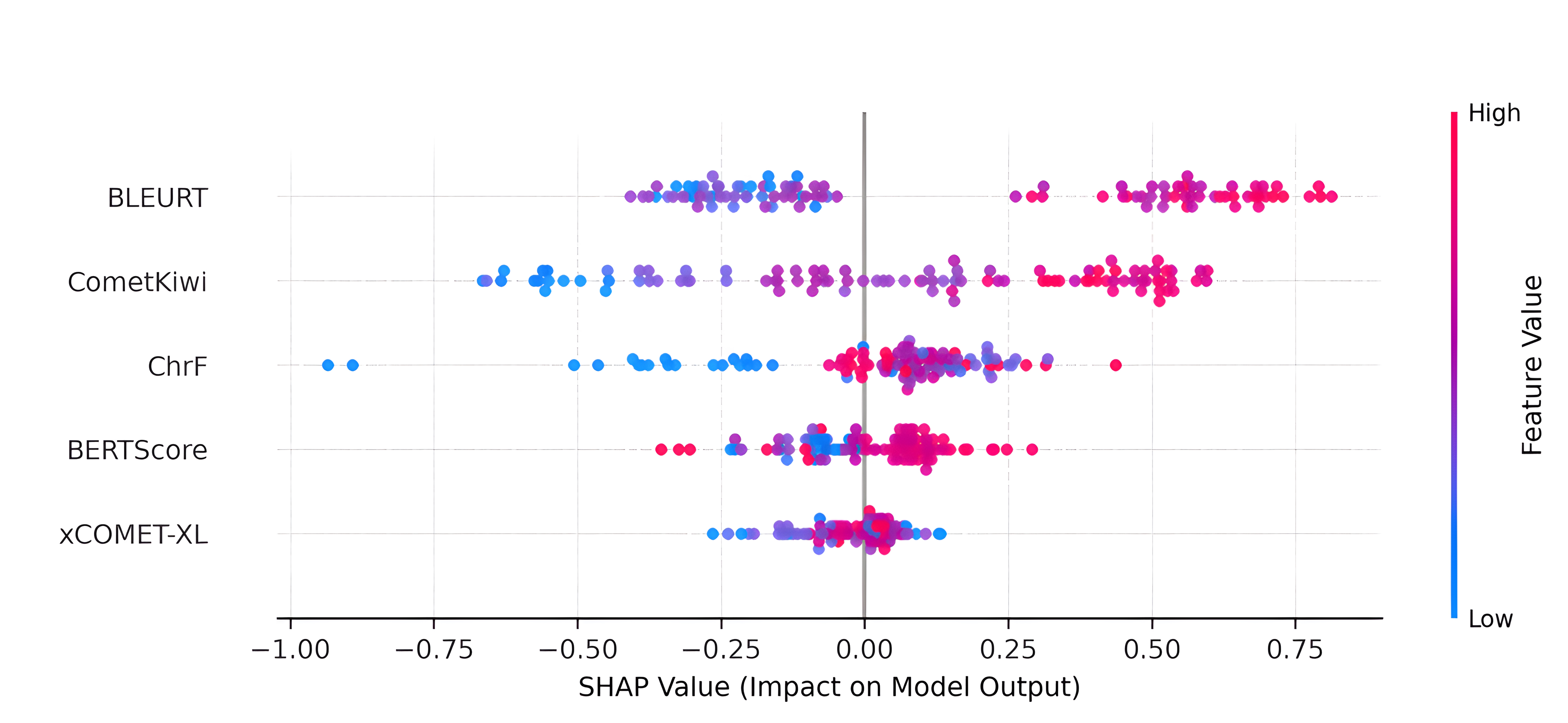}
    \includegraphics[width=0.28\linewidth]{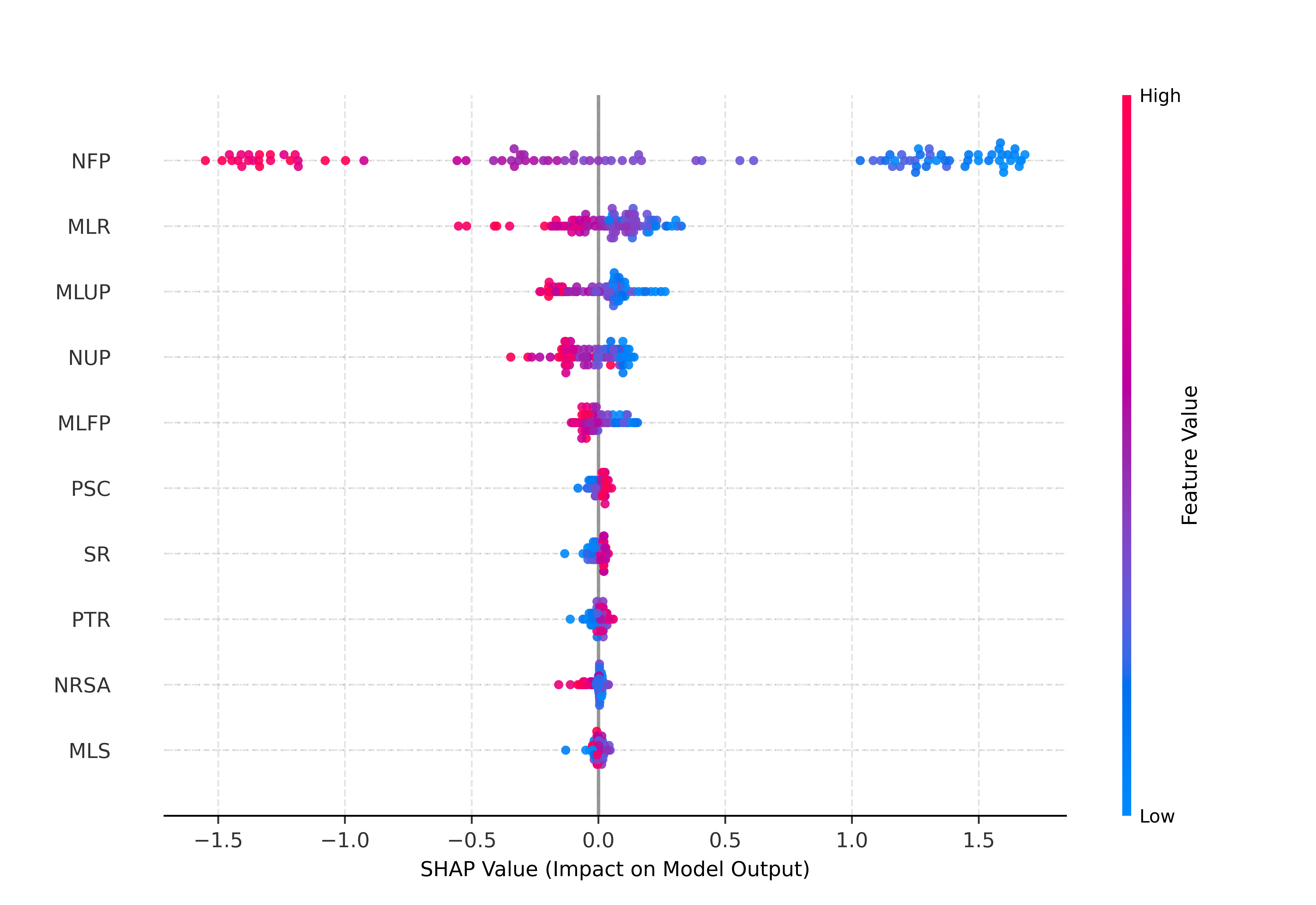}
    \includegraphics[width=0.28\linewidth]{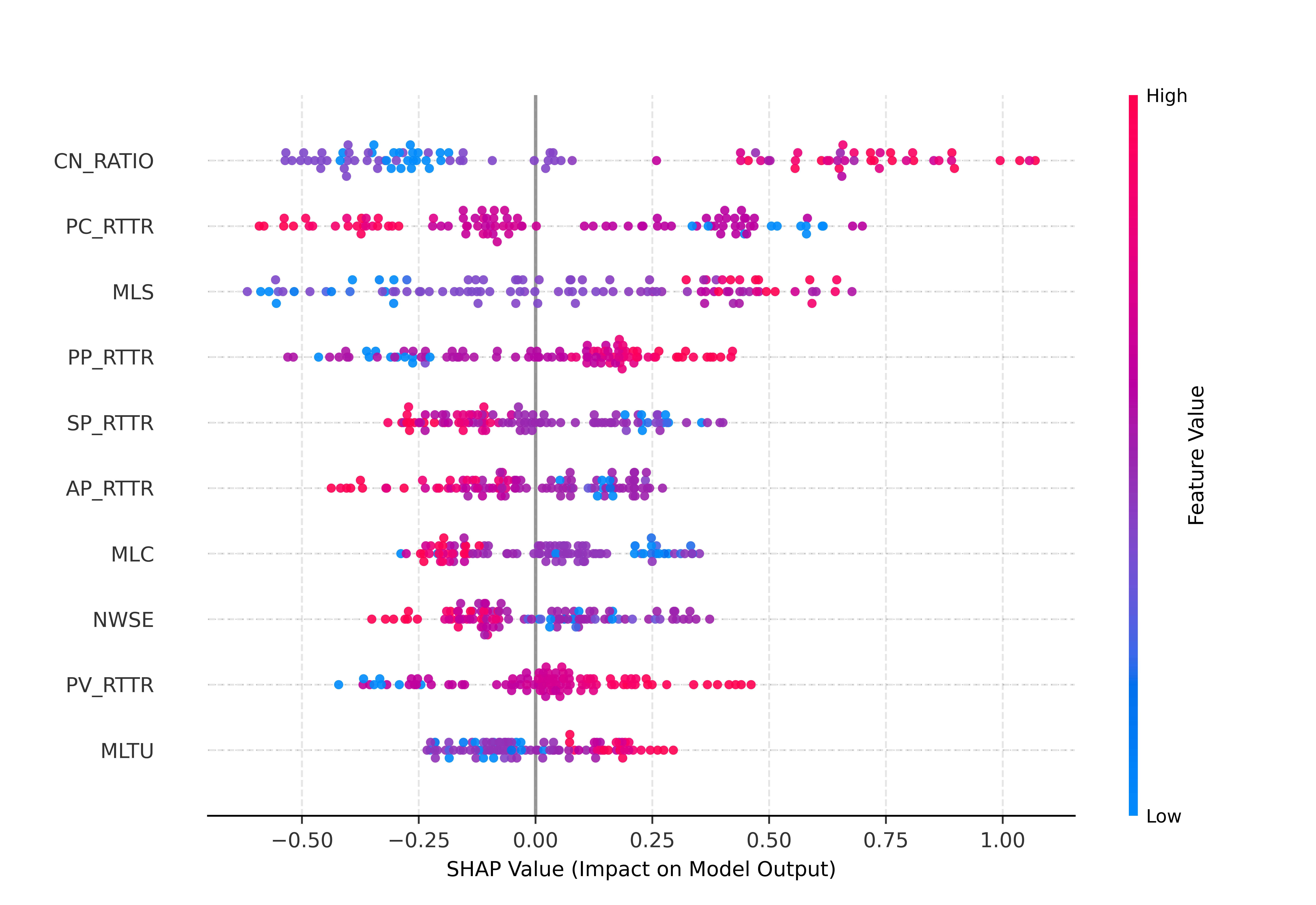}
    \caption{SHAP-Based global feature importance for InfoCom (left), FluDel (middle), and TLQual (right) predictions. Warmer tones (e.g., red) signify higher feature values and cooler tones (e.g., blue) indicate lower feature values. The features are arranged in descending order along the y-axis based on their global importance. The meaning of FluDel and TLQual features are given in Table~\ref{tab:FluDel-features}, \ref{tab:TLQual-features} respectively.}
    \label{fig:importance}
\end{figure*}

The inherent limitations of human evaluation have spurred considerable interest in automated assessment. However, existing works are characterized by both a thematic imbalance and methodological constraints. \textbf{Among the three established dimensions of interpreting quality (fidelity, fluency, and language use), investigations have disproportionately focused on the first two}, while language use has received scant scholarly attention~\citep{Yu-Heuven2017,Han-Yang2023,Wang-Wang2024,Han-Lu2021,Lu-Han2023}. Furthermore, prior research has predominantly relied on conventional statistical methods such as correlation and regression analyses~\citep{Yu-Heuven2017,Wang-Wang2024,Han-Lu2021,Lu-Han2023}, which are based on assumptions of linearity that often do not hold in complex, real-world datasets.

The advent of machine learning (ML) algorithms and large language models (LLMs) presents novel opportunities to analyze complex data patterns that elude traditional statistical methods. Nevertheless, \textbf{a notable obstacle in their application is the severe imbalance in data composition}. \citet{Wang-Yuan2023}, for example, find their five-class classification model unable to identify performances at the distributional extremes (``very poor'' and ``very good''), a direct consequence of the imbalanced training data distribution. \textbf{Another limitation is the inherent opacity of automated scoring systems}. \citet{Jia-Aryadoust2023}, for instance, find moderate correlations between GPT-4's interpreting performance assessment and human-assigned scores. Crucially, the internal decision-making processes of the LLMs remained opaque, with only the final scores being accessible. This ``black box'' nature severely restricts the diagnostic and educational utility of LLM scores.

In response to these challenges, we raise the following questions in this work:

1) Can we mitigate the underperformance of interpreting assessment models with data augmentation?


2) Which specific features of fidelity, fluency, and language use exhibit the strongest predictive power in interpreting assessment models?

3) What specific feature combinations influence individual student scores for each dimension of interpreting quality?

To answer these questions, we introduce a novel approach that combines feature engineering, data augmentation, and explainable AI (XAI) techniques~\citep{Arrieta-etal2019,Linardatos-etal2021} to evaluate interpreting performance across three key dimensions: fidelity, fluency, and target language quality. After using Variational Auto-Encoders (VAEs) to augment the data, we extract a broad set of features including translation quality metrics, temporal measures, and syntactic complexity indices to predict interpreting performance. Based on these features, we predict performance separately for each of the three dimensions adopting a multi-dimensional modeling strategy, which facilitates a more fine-grained analysis of interpreting quality and provides clearer insights into the specific contributions of features to each criterion. Furthermore, we apply Shapley Value (SHAP) analysis to provide interpretable explanations at both global and individual levels. To the best of our knowledge, we represent the first systematic efforts to automate the assessment of target language quality in interpreting.



\section{Related Work}
\subsection{Automated Interpreting Assessment}

The field of automated interpreting assessment is witnessing a paradigm shift, moving from statistical methods toward more sophisticated neural models. To date, the application of ML to interpreting quality evaluation remains a nascent but growing domain. The pioneering work by \citet{Le2016} developed estimators based on features from automatic speech recognition (ASR) and machine translation (MT), finding that MT features are most influential in predicting interpretation quality. Following that, \citet{Stewart2018} adapted the QuEst++ quality estimation pipeline with Support Vector Regression to predict the performance of simultaneous interpreters. More recently, \citet{Wang-Yuan2023} employed SVM and KNN algorithms to classify E-C interpretations, while \citet{Han-etal2025} further advanced the domain by integrating neural-based metrics with acoustic and linguistic indices through ordinal logistic regression.

\subsection{Dimensions of Interpreting Assessment}
\paragraph{Information Completeness}
Information completeness, also known as fidelity, refers to the extent of informational, semantic, and pragmatic correspondence between a source message and its translation~\citep{Han2017}. Existing metrics for automatic fidelity assessment can be broadly categorized into two types: non-neural and neural-based.

Non-neural metrics such as BLEU~\citep{Papineni2002} and chrF~\citep{Popovic2015} mainly rely on statistical and lexical matching to quantify the overlap of word or character sequences between a candidate translation and a human reference. Although these metrics have been widely adopted in the past decades, they have also been criticized for their reliance on surface-level comparisons that may not capture deeper semantic equivalence~\citep{Castilho2018}.

In contrast, neural-based metrics are derived from pre-trained language models and transcend surface matching by comparing contextualized embeddings. Prominent examples include BERTScore~\citep{Zhang2019}, BLEURT~\citep{Sellam2020}, CometKiwi~\citep{Rei2022}, and xCOMET~\citep{Guerreiro2024}. While \citet{Han-Lu2025} report a strong aggregate correlation between these scores and human evaluations on E-C interpreting, \citet{Lu-Han2023} find that the non-neural metrics BLEU and NIST outperform BERTScore, suggesting that non-neural and neural metrics may capture distinct, and potentially complementary, facets of interpreting quality.

\paragraph{Fluency}
Fluency is another key dimension of interpreting quality, reflecting how effectively and naturally an interpretation is delivered~\citep{Stenzl1983}. In computational modeling, fluency features are typically classified into three categories~\citep{Tavakoli2005}: (1) speed fluency, which captures the rate and density of delivery; (2) breakdown fluency, which measures speech continuity through the absence of interruptions and pauses; and (3) repair fluency, which quantifies self-corrections and repetitions. 

Within interpreting empirical research, considerable evidence has underscored the high predictive power of speed fluency features such as speech rate, phonation time ratio, and articulation rate~\citep{Han-Yang2023,Han2015,Song2020,Yu-Heuven2017}, while other works have also identified breakdown fluency features (e.g. mean length of unfilled pauses) as strong predictors~\citep{Wang-Wang2024,Wu2021}. In contrast, repair fluency features are less commonly employed and seldom show strong predictive effectiveness~\citep{Han2015}.


\paragraph{Target Language Use}
In interpreting assessment, target language quality typically refers to the grammaticality and idiomaticity of the target language output~\citep{Han2017}. Automated assessment of this dimension is facilitated by advances in computational tools such as Coh-Metrix~\citep{Graesser2004}, TAASSC~\citep{Kyle2016}, L2SCA~\citep{Lu2010}, and CCA~\citep{Hu2022a,Hu2022b}, which operationalize linguistic quality by calculating a wide array of features from lexical and phraseological indices to measures of syntax and discourse. While these features have been extensively applied in L2 writing and speaking research~\citep{Lu2010,Kyle-Crossley2017,Chen-etal2018}, their application to translation and interpreting contexts remains nascent, though existing findings show considerable promise~\citep{Ouyang2021,Han-etal2025,Han-etal2022}.

Yet, two key challenges remain. The first is the \textbf{need for more fine-grained feature design and application}. While coarse-grained metrics like T-unit complexity have long been valued~\citep{Ortega2003}, recent research advocates for supplementing them with fine-grained, usage-based indices that can capture subtle structural variations and better predict language development~\citep{Norris2009,Kyle-Crossley2017}. The second challenge concerns \textbf{language specificity}, as most NLP tools are developed primarily for English and may not fully account for the linguistic characteristics of other languages, such as the lack of overt morphological inflections and unique phraseological constructions in Chinese~\citep{Li1989,Hu2022a}

This evolving landscape is further complicated by the advent of LLMs. A recent large-scale study by \citet{mela} demonstrates that GPT-4o achieves near-human accuracy in grammatical acceptability judgment, leading to questions in the optimal combination of analytical tools — from established linguistic indices to emergent LLM-based judgments — that offers the most robust predictive power for assessing language use in interpreting.

\subsection{Data Augmentation in Interpreting Assessment}
Despite the aforementioned results in automatic interpreting assessment, the empirical application of ML is hampered by two fundamental and interrelated data challenges in this domain: \textbf{small sample size and imbalanced data composition}. The field is largely characterized by studies that rely on small datasets~\citep{Yu-Heuven2017,Lu-Han2023,Wang-Yuan2023,Wang-Wang2024}, substantially increasing the risk of overfitting. This problem is further exacerbated by pronounced class imbalance, as most datasets are heavily skewed toward average performance, with markedly fewer samples representing either very high or very low quality~\citep{Wang-Yuan2023,Han-etal2025}. 

To surmount these obstacles, data augmentation has emerged as a critical methodological intervention capable of enhancing model robustness and validity~\citep{Mumuni2022}. Common augmentation approaches include perturbation-based methods (adding Gaussian noise), interpolation techniques like SMOTE~\citep{SMOTE2002}, and generative models such as Generative Adversarial Networks (GANs) and Variational Autoencoders (VAEs)~\citep{Mumuni2022}. Among these, VAE offers three key advantages for ML-based interpreting assessment~\citep{Kingma2014}. First, its probabilistic framework captures complex interdependencies within fidelity, fluency, and language use features. Second, the continuous latent space enables smooth interpolation between existing samples to create coherent variations. Third, VAE preserves feature-label correspondence (i.e., the direct link between each sample's features and its corresponding interpreting quality score), which is crucial for maintaining assessment validity. \citet{Zhang2024} also demostrate the empirical viability of this technique.

\subsection{Explainable AI (XAI) and Its Application in Educational Contexts}
As educational AI systems become more sophisticated, XAI techniques are essential for understanding and validating these systems, thereby ensuring reliability, trust, and fairness~\citep{Gilpin2018,Rudin2019}.

Current XAI techniques fall into two main categories: intrinsic and post-hoc approaches~\citep{Gilpin2018,Rudin2019,Arrieta-etal2019,Linardatos-etal2021}. Intrinsic methods prioritize inherent interpretability by using transparent model architectures such as rule-based systems, decision trees, and linear models where coefficients directly indicate feature influence. In contrast, post-hoc methods explain already-trained black-box models without altering their structure, providing insights into complex models that would otherwise remain opaque. Most popular post-hoc methods - such as SHAP~\citep{Lundberg2017} and LIME~\citep{LIME2016} - provide feature attribution, while other methods also exist for example-based explanations and counterfactual explanations~\citep{Arrieta-etal2019,Linardatos-etal2021}. Based on their scope, post-hoc methods can also be categorized as either global explanation (illuminating overall model behavior across all instances) or local explanation (clarifying individual predictions).



For XAI research in education, learning analytics represents the most substantial area~\citep{Parkavi2024,Balachandar2025}, while applications have also been seen in automated language assessment, with most studies concentrating on explaining factors influencing performance quality~\citep{Kumar2020,Tang2024}. To our knowledge, \citet{Wang2024} is the only existing work to focus on explainability in automated interpreting assessment, which classifies interpreting quality into 5 levels and provides global explanations of feature importance using correlation analysis.


\section{Method}
\begin{figure}
    \centering
    \includegraphics[width=1\linewidth]{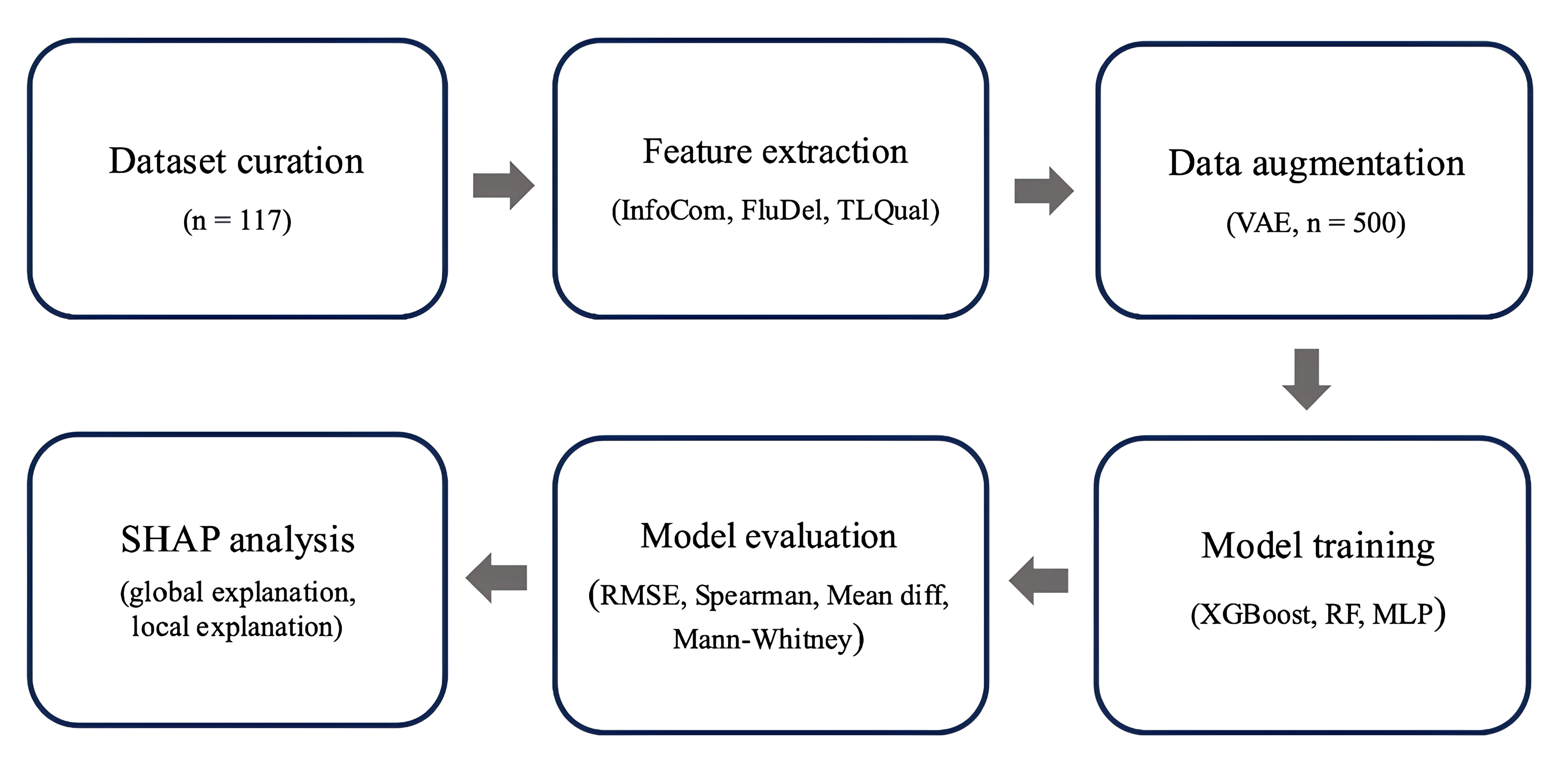}
    \caption{Methodological workflow of this study.}
    \label{fig:workflow}
\end{figure}

As illustrated in Figure~\ref{fig:workflow}, this study follows a structured method. First, we compile a new dataset comprising 117 student interpreting recordings in the English-Chinese direction, from which a range of linguistically meaningful and theoretically motivated features are extracted. To address challenges related to the small sample size and imbalanced score distribution, we employ VAE to generate new, realistic samples~\citep{Kingma2014}. After that, several machine learning models are trained to predict interpreting quality scores across different dimensions. Finally, to explain the inner decision-making of the trained models, we conduct a series of SHAP analyses.

\subsection{Original Dataset}
We compile a new dataset of 117 English-Chinese consecutive interpreting samples, collected from 39 undergraduate English majors at a university in Shanghai, China (Mean age = 18.47 years, SD = 1.13 years). All participants, whose L1 is Chinese and L2 is English, have passed CET-4 (College English Test), demonstrating satisfactory English proficiency. Before data collection, they completed 16 weeks (32 credit hours) of interpreting training.

The interpreting task uses six passages adapted from authentic public speeches, each containing an equal number of sentences and controlled for sentence length (M = 18.14 words, SD = 0.78 word). Further details about the passages, along with extracted linguistic feature values, are presented in Appendix~\ref{appendix-data}. These texts are converted into audio format using ElevenLabs' text-to-speech technology\footnote{\url{https://elevenlabs.io/}}. The resulting audio files feature standard pronunciation and averaged approximately 2 minutes in duration.

Assessment of the interpreting samples is conducted by three experienced raters, each with over three years of university-level teaching experience in domestic or international settings. The evaluation employes \citet{Han2017}'s four-band, eight-point analytic rubric, which assesses the three key dimensions of interpreting quality: InfoCom (information completeness), FluDel (fluency), and TLQual (target language quality). To ensure scoring consistency, raters underwent comprehensive training before the formal assessment. Detailed descriptions of the rater training procedures, student separation reliability, and the infit and outfit mean square statistics for each rater are provided in Appendix~\ref{appendix-training}. To mitigate potential inconsistencies and rater bias, Many-Facet Rasch Measurement (MFRM) analysis~\citep{Linacre2002} is used to calibrate raw scores and establish the final ground truth scores.

\subsection{Audio Processing}
To process the audio recordings of interpreting, we first use iFLYTEK ASR system\footnote{\url{https://global.xfyun.cn/products/real-time-asr}} to transcribe them into texts. To enhance annotation reliability, we implement a two-stage error detection process.

In the first stage, GPT-4o\footnote{GPT-4o-2024-08-06 with temperature set to 0.} is used for grammatical error diagnosis by adapting the framework of \citet{Rao2020} and \citet{Fu2018}. A structured prompt template (see Appendix~\ref{appendix-prompt}) is designed to guide GPT-4o's annotations, providing explicit instructions on four major error types: Redundant Words (R), Missing Words (M), Word Selection Errors (S), and Word Ordering Errors (W). To enhance the model's performance and reliability, we provide the it with few-shot examples and instructe it to explicitly articulate its decision-making process for each identified error and provide a corresponding confidence level. Particularly, we specify in the guidelines that filled pauses (e.g., ``uh'') are not considered errors, and analysis should focus solely on the final sentence version, disregarding repetitions, false starts, or self-corrections.

In the second stage, each transcription is manually reviewed and corrected. We recruited two postgraduate students in linguistics to independently annotate 100 randomly selected sentences, following the same guidelines as GPT. Inter-annotator agreement among human annotators yields a Cohen's Kappa coefficient of 0.86, while agreement between GPT-4o annotations and human annotations achieves a Fleiss' Kappa coefficient of 0.71, indicating a substantial level of consistency.

\begin{table*}[ht]
    \centering
    \adjustbox{width=0.9\textwidth,center}{
    \rowcolors{2}{gray!10}{white}
    \begin{tabular}{lm{13cm}}
\toprule
\textbf{Feature} & \textbf{Short description} \\
\midrule
chrF & Measures n-gram overlap between the interpreted and reference text \\
BLEURT-20 & Assesses the semantic similarity between the interpreted text and reference text based on contextualized embeddings from BERT and RemBERT \\
BERTScore & Measures the similarity between interpreted and reference translations by computing cosine similarity of their contextualized embeddings using BERT \\
CometKiwi-da & A reference-free regression model based on the InfoXLM architecture, trained on direct assessments from WMT17-WMT20 and the MLQE-PE corpus \\
xCOMET-XL & An extension of COMET, designed to identify error spans and assign quality scores, achieving state-of-the-art correlation with MQM error typology-derived scores \\
\bottomrule
    \end{tabular}
    }
    \caption{Features adopted for InfoCom assessment.}
    \label{tab:infocom-features}
\end{table*}

\begin{table*}[ht!]
    \centering
    \adjustbox{width=0.9\textwidth,center}{
    \rowcolors{2}{gray!10}{white}
    \begin{tabular}{lm{3.5cm}m{11cm}}
\toprule
\textbf{Feature} & \textbf{Full Name} & \textbf{Description} \\
\midrule
SR & Speech Rate & The overall pace of speech, calculated as the number of syllables uttered per second. \\
AR & Articulation Rate & The rate of syllable production, excluding pauses. \\
PTR & Phonation Time Ratio & The proportion of time spent vocalizing relative to the total duration. \\
MLS & Mean Length of Syllables & The average duration of each syllable. \\
MLR & Mean Length of Run & The average number of syllables produced in a continuous stream. \\
PSC & Pruned Syllable Count & The total syllable count after removing filled pauses. \\
NFP & Number of Filled Pauses & The frequency of filled pauses (e.g., ``um,'' ``uh''). \\
NUP & Normalized Number of Unfilled Pauses & The frequency of silent pauses. An unfilled pause is defined as a silence of 0.35 seconds or longer, consistent with recommendations for E-C interpreting~\citep{Mead2005}. \\
MLFP & Mean Length of Filled Pauses & The average duration of filled pauses. \\
MLUP & Mean Length of Unfilled Pauses & The average duration of silent pauses. \\
NRLFP & Number of Relatively Long Filled Pauses & The number of filled pauses longer than Q3 + 1.5 * Interquartile Range (IQR) and shorter than or equal to Q3 + 3 * IQR. \\
NRLUP & Number of Relatively Long Unfilled Pauses & The number of unfilled pauses longer than Q3 + 1.5 * IQR and shorter than or equal to Q3 + 3 * IQR. \\
NRSA & Number of Relatively Slow Articulations & The number of syllables longer than Q3 + 1.5 * IQR and shorter than or equal to Q3 + 3 * IQR. \\
NPSA & Number of Particularly Slow Articulations & The number of syllables longer than Q3 + 3 * IQR. \\
\bottomrule
    \end{tabular}
    }
    \caption{14 FluDel features examined in this work.}
    \label{tab:FluDel-features}
\end{table*}

\begin{table*}[ht]
    \centering
    \adjustbox{width=0.9\textwidth,center}{
    \rowcolors{2}{gray!10}{white}
    \begin{tabular}{m{5cm}m{5.5cm}m{5.5cm}}
\toprule
Coarse-Grained & Phraseological Diversity & Phraseological complexity \\
\midrule
Mean Length of Sentences (MLS) & Verb-Object Root Type-Token Ratio (VO\_RTTR) & Verb-Object Combination Ratio (VO\_RATIO) \\
Mean Length of T-units (MLTU) & Subject-Predicate Root Type-Token Ratio (SP\_RTTR) & Subject-Predicate Combination Ratio (SP\_RATIO) \\
Number of T-units Per Sentence (NTPS) & Adjective-Noun Root Type-Token Ratio (AN\_RTTR) & Adjective-Noun Combination Ratio (AN\_RATIO) \\
Mean Length of Clauses (MLC) & Adverb-Preposition Root Type-Token Ratio (AP\_RTTR) & Adverb-Preposition Combination Ratio (AP\_RATIO) \\
Number of Clauses Per Sentence (NCPS) & Classifier-Noun Root Type-Token Ratio (CN\_RTTR) & Classifier-Noun Combination Ratio (CN\_RATIO) \\
  & Preposition-Postposition Root Type-Token Ratio (PP\_RTTR) & Preposition-Postposition Combination Ratio (PP\_RATIO) \\
  & Preposition-Verb Root Type-Token Ratio (PV\_RTTR) & Preposition-Verb Combination Ratio (PV\_RATIO) \\
  & Predicate-Complement Root Type-Token Ratio (PC\_RTTR) & Predicate-Complement Combination Ratio (PC\_RATIO) \\
\bottomrule
    \end{tabular}
    }
    \caption{21 Syntactic complexity features adopted for TLQual assessment.}
    \label{tab:TLQual-features}
\end{table*}

\subsection{Feature Extraction}\label{sec:method-feature-extraction}
Each scoring dimension of interpreting is represented by a distinct set of extracted features. Fluency features are extracted from the original transcript, while other features are derived from cleaned transcripts (after removing fillers, false starts, and self-repair).

For \textbf{InfoCom}, we use five established metrics from the field of machine translation quality assessment to measure the preservation of information from source to target language (Table~\ref{tab:infocom-features}).

\textbf{FluDel} features include 14 temporal features~(Table~\ref{tab:FluDel-features}) derived from prior research~\citep{Barik1973,Yu-Heuven2017,Song2020,Wang-Wang2024}. These features can be categorized into two groups: speed fluency features (1–6) and breakdown fluency features (7–14). Features related to unfilled pauses are extracted automatically using Python packages librosa (v0.10.2) and soundfile (v0.12.1), with pauses identified based on an intensity threshold of -18 dB as recommended by \citet{Wu2021}. Additional features are derived from time-aligned transcriptions generated by the iFLYTEK ASR system.

\textbf{TLQual} is evaluated through 25 features related to syntactic complexity and grammatical accuracy. Among them, 21 syntactic complexity features encompassing both coarse-grained and fine-grained measures (Table~\ref{tab:TLQual-features}) are extracted using Chinese Collocation Analyzer (CCA, \citealp{Hu2022a,Hu2022b}), which is specifically developed for L2 Chinese texts, making it particularly appropriate for E-C interpreting studies. The remaining 4 grammatical accuracy features are derived from the grammatical error annotations by GPT-4o, specifically Number of Redundant Words (NRW), Number of Missing Words (NMW), Number of Word Selection Errors (NWSE), and Number of Word Ordering Errors (NWOE).

\subsection{Data Augmentation}
Unlike general L2 learners, interpreting students constitute a smaller pool due to the advanced linguistic competence and cognitive demands required by the task. This scarcity underscores the need for data augmentation techniques to increase the quantity and diversity of learner datasets~\citep{Mumuni2022}.
	
In line with the approach proposed by \citet{Zhang2024}, we employ Variational Autoencoder (VAE) to address the challenge of score distribution imbalance in the original dataset. The primary objective is to generate realistic, synthetic feature vectors for the three distinct dimensions of interpreting quality being assessed. To achieve this, we train a separate conditional VAE for each of the three dimensions.
The synthetic feature vectors generated by these VAE models are then combined with the original 117 data points, resulting in an augmented dataset comprising 500 samples.

\subsection{Model Training and Validation}
Three types of machine learning models — XGBoost, Random Forest (RF), and Multi-Layer Perceptron (MLP) — are employed to predict the InfoCom, FluDel, and TLQual scores. The modeling process followes a systematic procedure that consists of feature extraction, feature standardization, data splitting, model training and validation, and model testing~\citep{Mienye2022}.

All extracted features (as detailed in Section~\ref{sec:method-feature-extraction}) are first standardized using z-score normalization. The initial dataset is then split into training (80\%) and testing (20\%) subsets. Following the data split, model training and validation are conducted with five-fold cross-validation and a grid search for hyperparameters, using root mean square error (RMSE) as validation criterion.

After cross-validation and hyperparameter optimization, the best-performing configuration is selected for each model. Each final model is then retrained on the entire training set using the optimal hyperparameters and subsequently evaluated on the held-out test set to assess its predictive performance on unseen data. Multiple evaluation metrics are employed in this stage to provide a comprehensive assessment of model quality, including:

(1) RMSE: measures the magnitude of prediction errors.

(2) Spearman's ($\rho$): assesses the monotonic relationship between predicted and actual scores.

(3) Mean absolute error (MAE): quantifies the average absolute deviation between predicted and actual scores, providing a direct measure of prediction accuracy.

(4) Mann-Whitney U Test: determines whether there are significant differences in the distributions of predicted and actual scores.

(5) Exact Agreement rate (EAR): quantifies the proportion of predictions that exactly match the actual scores after both are rounded to the nearest integer. Rounding is required because our models predict continuous MFRM-calibrated scores (1-8), and agreement is typically assessed against discrete levels.

(6) Adjacent Agreement rate (AAR): measures the proportion of predictions that fall within one integer unit (either +1 or -1) of the actual scores after both are rounded to the nearest integer.

Beyond these overall metric values, we also perform case studies of prediction errors to gain more in-depth insights into specific aspects of the model's performance.

\subsection{Result Explanation Using XAI Techniques}
We further employ SHAP to interpret model behavior at two levels: the overall model (global explanations) and individual predictions (local explanations). Global explanations offer a broader perspective by summarizing the overall impact of features across the entire dataset. Local explanations, on the other hand, provide insights into how individual features influence a single predicted outcome. These analyses are implemented using the shap library\footnote{\url{https://shap.readthedocs.io/en/latest/index.html}}.
\begin{figure*}[ht]
    \centering
    \includegraphics[width=0.85\linewidth]{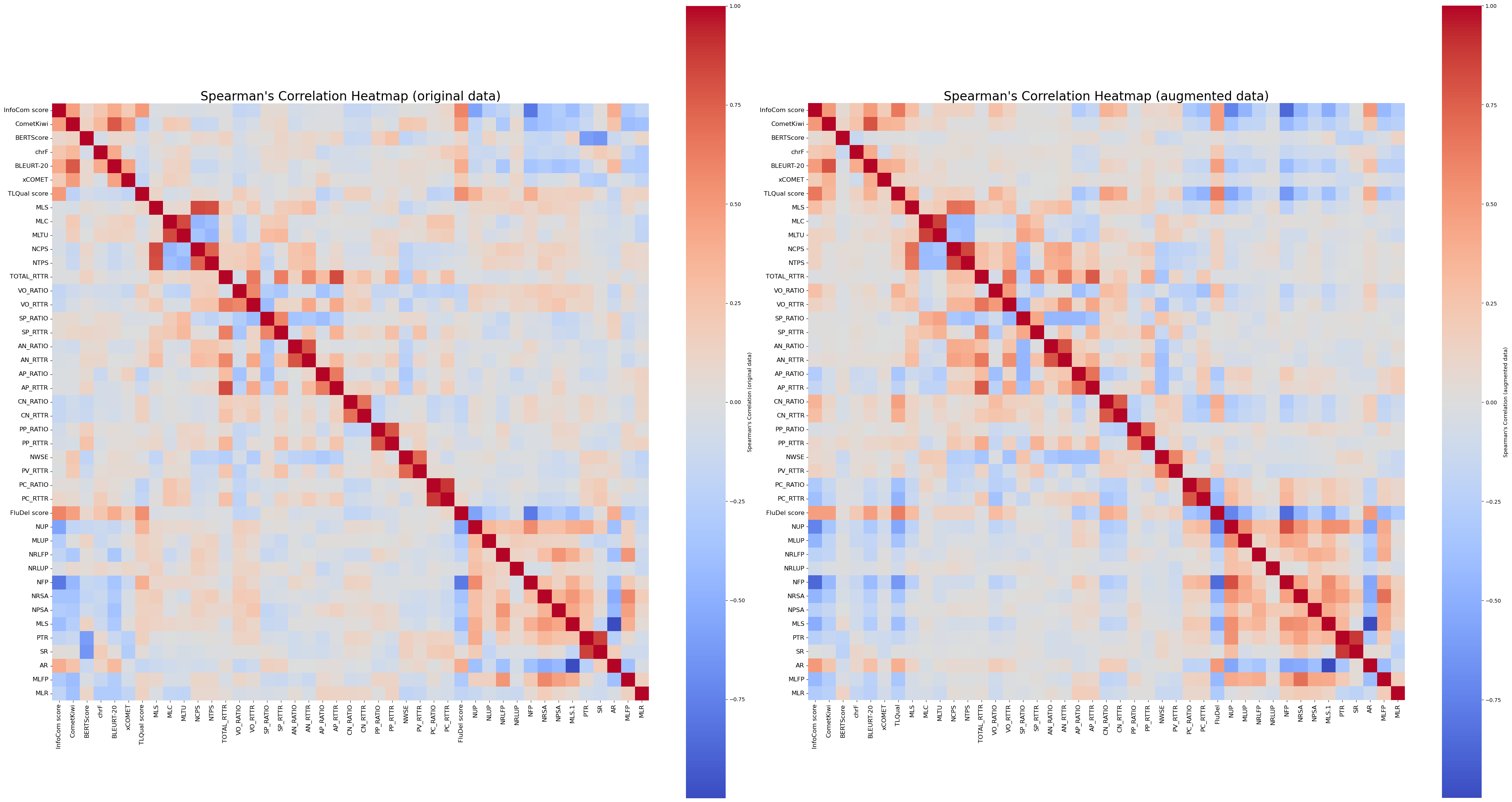}
    \caption{Pairwise correlation heatmap between features and scores.}
    \label{fig:feature_score_correlation}
\end{figure*}

\section{Results}
\subsection{Descriptive Statistics}
\begin{figure}[ht]
    \centering
    \includegraphics[width=1\linewidth]{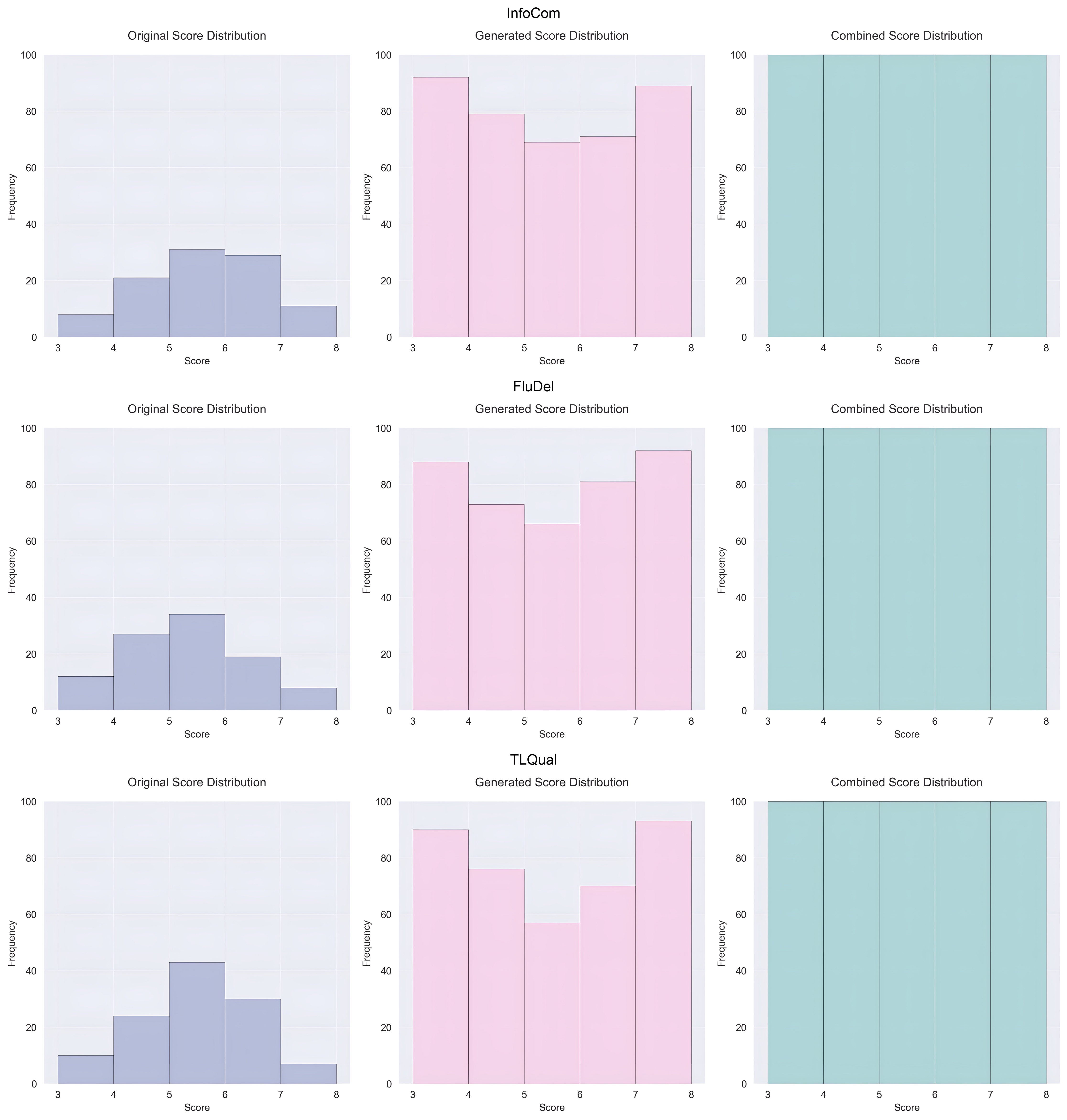}
    \caption{Distribution of raw (left), generated (middle), and augmented (right) data.}
    \label{fig:data_distribution}
\end{figure}

Due to the consistently reasonable level of interpreting proficiency demonstrated by all student participants, the dataset lacks samples with scores in the 1-2 range. However, Figure~\ref{fig:data_distribution} demonstrates that data augmentation has successfully achieved an approximately uniform distribution of interpretation scores on the remaining range. Table~\ref{tab:data-statistics} further reveals that compared with the original data, the augmented data exhibits very close mean values and marginally insreased standard deviations in all three dimensions. The descriptive statistics for all features used in this work are provided in Appendix~\ref{appendix-features}, and the pairwise Spearman's correlations between features and scores in both the original and augmented datasets are illustrated in Figure~\ref{fig:feature_score_correlation}.

\begin{table}[th]
    \centering
    \adjustbox{width=\linewidth,center}{
    \begin{tabular}{llcccc}
\toprule
Score &  & Mean & SD & Skewness & Kurtosis \\
\midrule
\multirow{2}{*}{InfoCom} & Raw & 5.32 & 1.35 & -0.37 & 2.25 \\
 & Aug. & 5.33 & 1.47 & -0.05 & -0.51 \\
\multirow{2}{*}{FluDel} &\cellcolor{gray!10} Raw &\cellcolor{gray!10} 4.93 &\cellcolor{gray!10} 0.77 &\cellcolor{gray!10} -0.31 &\cellcolor{gray!10} 2.94 \\
 &\cellcolor{gray!10} Aug. &\cellcolor{gray!10} 4.95 &\cellcolor{gray!10} 0.98 &\cellcolor{gray!10} -0.10 &\cellcolor{gray!10} -0.67 \\
\multirow{2}{*}{TLQual} & Raw & 5.21 & 0.95 & -0.23 & 3.38 \\
 & Aug. & 5.24 & 1.06 & 0.06 & -0.85 \\
\bottomrule
    \end{tabular}
    }
    \caption{Descriptive statistics for scores from the raw data and augmented data.}
    \label{tab:data-statistics}
\end{table}

\begin{table*}[ht]
    \centering
    \begin{tabular}{lllrrrrrr}
\toprule
Score & Model & Data & RMSE & Spearman & MAE & Mann-Whitney U & EAR & AAR \\
\midrule
\multirow{6}{*}{InfoCom} & \multirow{2}{*}{XGBoost} & raw & 1.36 & 0.49$^{**}$ & 0.95 & 259 (p = 0.70) & 0.63 & 0.83 \\
 &  & aug. & 1.17 & 0.62$^{**}$ & 0.49 & 5751 (p = 0.12) & 0.71 & 0.86 \\
 & \multirow{2}{*}{RF} &\cellcolor{gray!10} raw &\cellcolor{gray!10} 1.42 &\cellcolor{gray!10} 0.51$^{**}$ &\cellcolor{gray!10} 0.87 &\cellcolor{gray!10} 209 (p = 0.45) &\cellcolor{gray!10} 0.67 &\cellcolor{gray!10} 0.88 \\
 &  &\cellcolor{gray!10} aug. &\cellcolor{gray!10} 1.05 &\cellcolor{gray!10} 0.68$^{**}$ &\cellcolor{gray!10} 0.41 &\cellcolor{gray!10} 5693 (p = 0.15) &\cellcolor{gray!10} 0.77 &\cellcolor{gray!10} 0.90 \\
 & \multirow{2}{*}{MLP} & raw & 2.43 & 0.43$^{*\phantom{*}}$ & 1.21 & 215 (p = 0.53) & 0.54 & 0.75 \\
 &  & aug. & 1.25 & 0.58$^{**}$ & 0.79 & 5744 (p = 0.12) & 0.68 & 0.77 \\
\midrule
\multirow{6}{*}{FluDel} & \multirow{2}{*}{XGBoost} & raw & 0.84 & 0.69$^{**}$ & 0.65 & 272 (p = 0.49) & 0.69 & 0.83 \\
 &  & aug. & 0.68 & 0.87$^{**}$ & 0.41 & 5375 (p = 0.36) & 0.72 & 0.91 \\
& \multirow{2}{*}{RF} &\cellcolor{gray!10} raw &\cellcolor{gray!10} 0.70 &\cellcolor{gray!10} 0.65$^{**}$ &\cellcolor{gray!10} 0.68 &\cellcolor{gray!10} 274 (p = 0.46) &\cellcolor{gray!10} 0.71 &\cellcolor{gray!10} 0.83 \\
 &  &\cellcolor{gray!10} aug. &\cellcolor{gray!10} 0.61 &\cellcolor{gray!10} 0.86$^{**}$ &\cellcolor{gray!10} 0.43 &\cellcolor{gray!10} 5302 (p = 0.46) &\cellcolor{gray!10} 0.75 &\cellcolor{gray!10} 0.93 \\
& \multirow{2}{*}{MLP} & raw & 1.74 & 0.39$^{**}$ & 1.17 & 274 (p = 0.46) & 0.54 & 0.71 \\
 &  & aug. & 1.20 & 0.53$^{**}$ & 0.89 & 4621 (p = 0.36) & 0.64 & 0.82 \\
\midrule
\multirow{6}{*}{TLQual} & \multirow{2}{*}{XGBoost} & raw & 0.87 & 0.66$^{**}$ & 0.72 & 267 (p = 0.41) & 0.67 & 0.83 \\
 &  & aug. & 0.75 & 0.79$^{**}$ & 0.45 & 5386 (p = 0.33) & 0.76 & 0.91 \\
& \multirow{2}{*}{RF} &\cellcolor{gray!10} raw &\cellcolor{gray!10} 0.97 &\cellcolor{gray!10} 0.58$^{**}$ &\cellcolor{gray!10} 0.86 &\cellcolor{gray!10} 232 (p = 0.42) &\cellcolor{gray!10} 0.63 &\cellcolor{gray!10} 0.79 \\
 &  &\cellcolor{gray!10} aug. &\cellcolor{gray!10} 0.92 &\cellcolor{gray!10} 0.73$^{**}$ &\cellcolor{gray!10} 0.54 &\cellcolor{gray!10} 5522 (p = 0.20) &\cellcolor{gray!10} 0.78 &\cellcolor{gray!10} 0.89 \\
 & \multirow{2}{*}{MLP} & raw & 1.58 & 0.45$^{*\phantom{*}}$ & 1.10 & 206 (p = 0.40) & 0.58 & 0.75 \\
 &  & aug. & 1.04 & 0.62$^{**}$ & 0.83 & 4973 (p = 0.95) & 0.69 & 0.85 \\
\bottomrule
    \end{tabular}
    \caption{Performance of machine learning regressors trained on raw and augmented data. $^{**}p<0.01$; $^{*}p<0.05$.}
    \label{tab:performance}
\end{table*}

\subsection{Effectiveness of Models Trained on Raw and Augmented Data}
As shown in Table~\ref{tab:performance}, XGBoost trained on the augmented dataset achieves the highest performance in predicting FluDel and TLQual scores, representing an improvement over its already robust performance on the raw dataset. For InfoCom prediction, the RF regressor trained on augmented data yields the best results, also substantially outperforming the same model trained on raw data. In contrast, MLP consistently exhibits the lowest performance, though also showing a notable improvement when trained on augmented data. In Appendix~\ref{appendix-case}, we provide detailed analyses of instances where model predictions diverge greatly from human scores, offering nuanced insights into the models' performance characteristics.

\subsection{Global explanations of model prediction}
Figure~\ref{fig:importance} (left) illustrates the global feature importance of the best-performing RF regressor for InfoCom score prediction. Among these, BLEURT (M = 0.32, 95\% CI\footnote{To assess the stability of feature contributions, a bootstrap procedure is conducted with 1,000 resamples drawn from the augmented dataset. For each bootstrap sample, SHAP values are computed using the best-performing ML model. The mean SHAP value for each feature is recorded across iterations to estimate its average effect on predictions. 95\% Confidence intervals (CI) are calculated as the 2.5th and 97.5th percentiles of the bootstrapped distribution, capturing both the direction and magnitude of each feature's influence.} = [0.25, 0.37]), CometKiwi (M = 0.17, 95\% CI = [0.08, 0.26]), and chrF (M = 0.07, 95\% CI = [0.04, 0.09]) demonstrate the highest mean SHAP values. In other words, higher values of these metrics are positively associated with higher predicted InfoCom scores.

As illustrated in Figure~\ref{fig:importance} (middle), NFP (M = -0.17, 95\% CI = [-0.27, -0.10]) exhibits the strongest negative effect on FluDel scores, with higher NFP values leading to lower predictions by the XGBoost regressor. Similarly, other breakdown fluency features, including MLUP, NUP, and MLFP also negatively impact predicted outcomes. Speed fluency features such as PSC, SR, PTR, and MLS have a positive but very small impact on the model's predictions, while MLR yields a negative effect instead.

Figure~\ref{fig:importance} (right) demonstrates that the grammatical accuracy index NWSE (M = -0.09, 95\% CI = [-0.15, -0.04]) has an inverse relationship with model predictions, indicating that a higher frequency of word selection errors corresponds to lower predicted scores. Among phraseological complexity features, CN\_RATIO (M = 0.25, 95\% CI = [0.18, 0.31]) has the most significant influence, with higher values leading to increased predictions. In addition, a group of phraseological diversity metrics also contribute positively to model output, including PP\_RTTR and PV\_RTTR. In contrast, AP\_RTTR and PC\_RTTR exhibit negative effects. For coarse-grained features, higher MLC values are associated with lower model predictions, while MLS positively influences predicted outcomes.

\begin{figure*}[ht]
    \centering
    \includegraphics[width=1\linewidth]{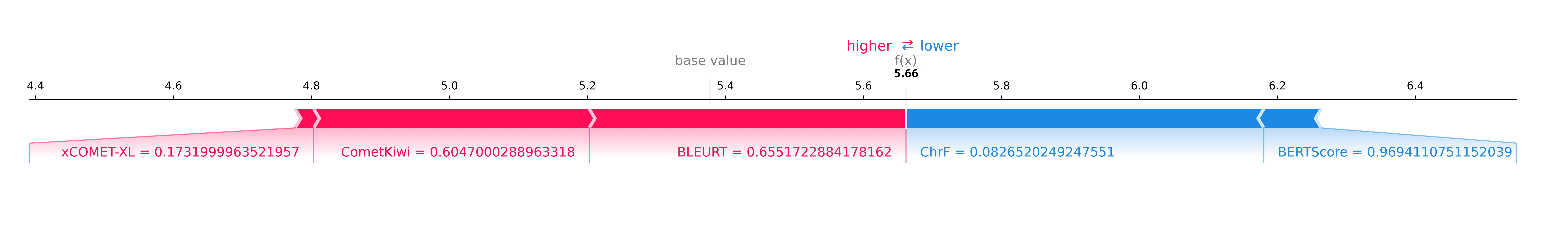}
    \caption{SHAP force plot for the InfoCom prediction of Sample 25.}
    \label{fig:forceplot_infocom_25}
\end{figure*}
\begin{figure}[ht]
    \centering
    \includegraphics[width=1\linewidth]{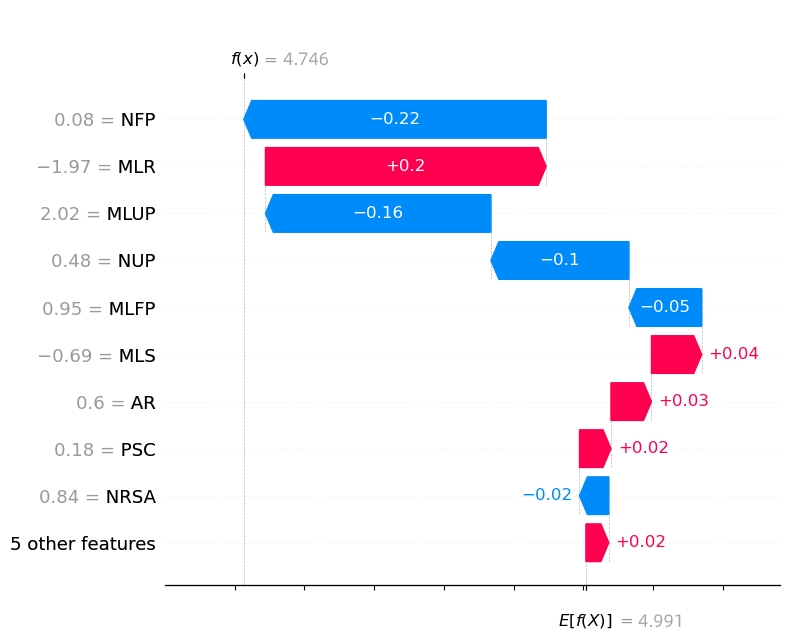}
    \caption{SHAP waterfall plot for the predicted FluDel score of Sample 50.}
    \label{fig:waterfullplot_fludel_50}
\end{figure}
\begin{figure}[ht]
    \centering
    \includegraphics[width=1\linewidth]{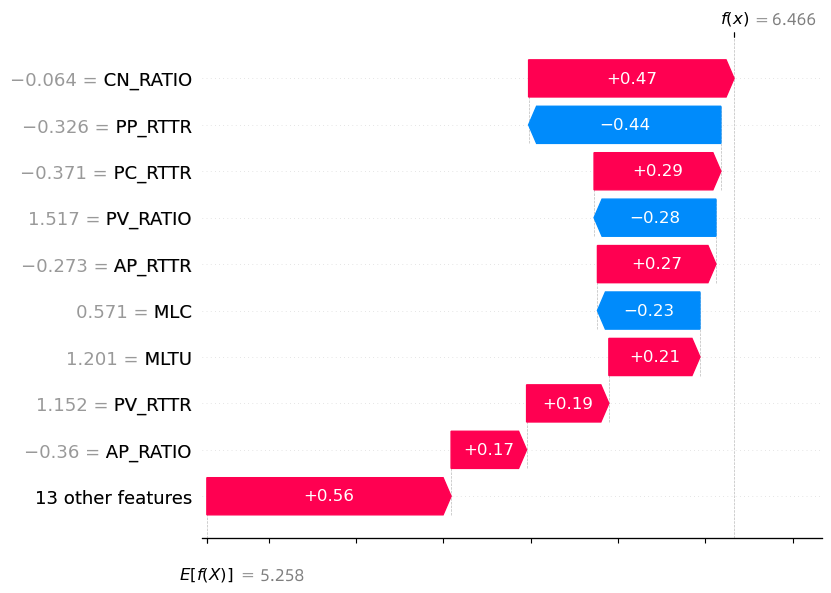}
    \caption{SHAP waterfall plot for the predicted TLQual score of Sample 87.}
    \label{fig:waterfallplot_tlqual_87}
\end{figure}

\subsection{Local explanations of model prediction}
Figure~\ref{fig:forceplot_infocom_25} illustrates the SHAP force plot for the InfoCom prediction of Sample 25, providing a detailed depiction of individual feature contributions. The plot is centered around the base value (approximately 5.4), representing the mean model output across the training dataset. The cumulative contributions of the InfoCom features slightly elevate the prediction to 5.66. Among these, BLEURT and COMET-Kiwi exert the most significant positive influence, whereas chrF contributes negatively. The relatively high BLEURT and COMET-Kiwi scores suggest that Sample 25 retains most of the source information, albeit with some loss, while the markedly low chrF score indicates substantial lexical and syntactic divergence from the reference text.

In Figure~\ref{fig:waterfullplot_fludel_50}, the SHAP waterfall plot for the FluDel prediction of Sample 50 is shown. The expected value E[f(x)] = 4.991 represents the mean model output across the training dataset. Feature contributions collectively reduce the prediction to f(x) = 4.746. Among these, the pause-related features - NFP, MLUP, and NUP - exhibit the most pronounced negative impact, decreasing the prediction by 0.22, 0.16, and 0.1, respectively. Conversely, MLR has the strongest positive effect, increasing the prediction by 0.2. These findings suggest that the interpreter may need to enhance pause management by minimizing both the frequency and duration of pauses while striving for more extended, uninterrupted speech production.

The SHAP waterfall plot for the TLQual prediction of Sample 87 is depicted in Figure~\ref{fig:waterfallplot_tlqual_87}. The model's expected value is E[f(x)] = 5.258, with feature contributions collectively increasing the prediction to 6.466. Among these, CN\_RATIO is the most influential positive factor, increasing the prediction by 0.47. Other contributing features include PC\_RTTR, AP\_RTTR, PV\_RTTR, and AP\_RATIO. Conversely, PP\_RTTR exerts the most significant negative influence, reducing the prediction by 0.44, with additional negative contributions from PV\_RATIO and MLC. These results indicate that the diversified and sophisticated use of CN, PC, AP, PV, and AP structures aligns with typical language patterns in this context. However, excessive use of PP structures (e.g. \begin{CJK*}{UTF8}{gkai}在...上, 当...时\end{CJK*}) appears detrimental. Additionally, the negative impact of MLC suggests that complex clauses could be restructured into simpler sentences or reformulated using a topic-comment structure, a common grammatical pattern in Chinese.

\section{Discussions}
\subsection{Modeling effectiveness and the impact of data augmentation}
Our analysis indicates that the selected machine learning algorithms demonstrate robust performance on the augmented dataset, with RF yielding the best results for InfoCom score estimation and XGBoost performing best on FluDel and TLQual. Comparing with previous models that only perform well on middle range scores but failing at lower and higher ranges~\citep{Han-etal2025,Wang-Yuan2023}, our results underscore the importance of data augmentation in improving model performance, particularly for predicting scores at extreme ends of the scale.


\subsection{Global explanations of feature importance}
\paragraph{Information completeness}
Our SHAP analysis identifies the two neural-based metrics, BLEURT and CometKiwi, as having the greatest influence on the global prediction of InfoCom scores, aligning with previous research by \citet{Han-Lu2025}. The superior performance of BLEURT is likely attributable to its extensive pre-training on synthetic data and its ability to incorporate diverse lexical and semantic signals, which enables the metric to capture more nuanced linguistic patterns compared to BERTScore~\citep{Han-Lu2025}. Conversely, the relatively low performance of XCOMET may stem from a misalignment between its training paradigm (error annotation) and the assessment context (analytical rubric scoring). 

\paragraph{Fluency of delivery}
Our findings reveal that NFP has the most pronounced negative impact on the model's global prediction of FluDel scores, followed by other pause-related features including MLUP, NUP, and MLFP, aligning with previous findings~\citep{Yu-Heuven2017}. In contrast, most speed fluency features (e.g. PSC, PTR, SR) exhibit small positive effects, although higher MLR values are linked to decreased predictions. We hypothesize that the negative role of MLR stems from the phenomenon that excessively long runs do not reflect controlled, fluent delivery but rather a form of ``run-on speech''. The interpreter, under high cognitive load, may be rushing to output information without strategic pausing for emphasis or listener comprehension~\citep{Lennon1990,Mead2005}, leading to human raters perceiving the speech as poorly managed and difficult to process.



\paragraph{Target language quality}
Among the GPT-4o-annotated features, NWSE exerts a significant negative effect on model predictions, underscoring the foundational role of grammatical accuracy in human judgments of language quality and mirroring findings in L2 speaking assessment~\citep{Li2024}.

Regarding length-related features, MLS has a positive effect on predictions, aligning with \citet{Zechner2017}. MLC, on the other hand, yields a negative impact, which is in sharp contrast to findings from other contexts such as L2 German and English speaking~\citep{NearySundquist2017,Bult2020}. This divergence likely stems from typological differences: the topic-comment structure of Chinese prioritizes discourse coherence, whereas the syntactic elaboration common in English relies more heavily on complex clausal dependencies~\citep{Li1989}. This suggests that in the Chinese interpreting context, longer but less syntactically dense sentences are perceived as higher quality.

Another key finding is the superior predictive importance of fine-grained features over coarse-grained ones. Within this category, features reflecting phraseological diversity (PC\_RTTR, PP\_RTTR, SP\_RTTR, AP\_RTTR, PV\_RTTR) are more influential than the single phraseological complexity feature (CN\_RATIO). Furthermore, our results reveal that Chinese-specific phraseological features (CN, PC, PP, PV) demonstrate greater importance than their language-independent counterparts (SP, AP). Taken together, these findings point towards the possibility that for E-C consecutive interpreting, a robust assessment of language use relies less on traditional measures of clausal complexity and more on the diverse and accurate use of language-specific phrasal units.

\subsection{The critical role of local explanations in automated interpreting assessment}
Local explanations in automated interpreting assessment offer significant value for both teaching and learning practices~\citep{Kumar2020,Tang2024,Gilpin2018,Rudin2019,Linardatos-etal2021}. For educators, these explanations provide actionable insights into the specific strengths and weaknesses of individual students' performances by highlighting the features that positively or negatively influence predicted scores. This enables teachers to tailor feedback and instructional strategies to target precise areas for improvement. For students, local explanations empower students to take ownership of their learning by focusing on specific performance aspects that require attention. 

Take the SHAP-based local explanation of FluDel prediction for Sample 50 as an example. Notably, pause-related features emerge as the primary detractors: NFP reduces the prediction by 0.22, MLUP by 0.16, and NUP by 0.1, indicating the student's difficulty with hesitation management. To address this, instructors can implement targeted exercises such as shadowing practices, where students reproduce source language with minimal delay~\citep{Christoffels2004}. The instructor could also implement targeted drills requiring students to deliver short segments without hesitation, progressively extending segment length while monitoring pause reduction. For reducing unfilled pauses specifically, anticipation exercises help students predict upcoming content elements, thereby decreasing processing latency~\citep{Chmiel2021}. Additionally, instructing in chunking strategies - organizing information into manageable units - can alleviate cognitive load that frequently manifests as extended pauses~\citep{Thalmann2019}.

In addition, the quantitative nature of SHAP values also allows instructors to prioritize interventions effectively. For this particular student, addressing filled pauses should take precedence over lengthy unfilled pauses, given its greater negative impact (0.22 vs. 0.16). Furthermore, tracking these SHAP contributions longitudinally across multiple performances enables instructors to monitor learning progression and intervention effectiveness, facilitating timely adjustments to teaching approaches as needed.
\section{Conclusion}
In this work, we propose an effective framework integrating feature engineering, ML models, data augmentation, and XAI for the multi-dimensional assessment of interpreting quality. A key finding is that VAE-based data augmentation substantially enhances model performance. Global XAI analysis reveales that fidelity prediction is most sensitive to neural-embedding metrics such as BLEURT, while fluency scores are primarily influenced by breakdown features, with NFP exerting the strongest negative effect. Target language quality, in turn, depends heavily on language-specific phraseological features, notably CN\_RATIO. These global insights are complemented by in-depth local explanations, which effectively diagnose individual strengths and weaknesses in performance. Looking forward, our method provides a promising direction in translating XAI-driven insights into pedagogical tools that deliver actionable feedback to trainees, thereby bridging the gap between automated assessment and student learning.

\bibliography{custom}

\clearpage
\appendix
\onecolumn
\section{More Details on Source Materials}\label{appendix-data}

\begin{table}[ht]
    \centering
    \adjustbox{width=\textwidth,center}{
    \begin{tabular}{llccccccc}
\toprule
Passage  &  Theme  &  DESWC  &  DESSL  &  DESWLlt  &  LDTTRa  &  RDFRE  &  RDFKGL  &  RDL2 \\
\midrule
1 &  Migration  & 185 & 19.32 & 5.11 & 0.72 & 35.25 & 15.05 & 13.37 \\
2 &  Migration  & 193 & 18.91 & 5.42 & 0.65 & 41.12 & 16.23 & 8.35 \\
3 &  Festival  & 182 & 19.75 & 5.16 & 0.75 & 29.74 & 16.51 & 8.90 \\
4 &  Festival  & 191 & 18.86 & 5.32 & 0.68 & 42.18 & 14.66 & 10.20 \\
5 &  Social equality  & 179 & 19.44 & 5.23 & 0.74 & 45.36 & 13.28 & 7.39 \\
6 &  Social equality  & 185 & 19.06 & 5.28 & 0.66 & 33.33 & 15.73 & 11.46 \\
\bottomrule
    \end{tabular}
    }
    \caption{Basic information on the six passages used in the interpreting tasks. DESWC: word count; DESSL: sentence length (number of words); DESWLlt: word length (mean); LDTTRa: lexical densitiy (type-token ratio); RDFRE: Flesch Reading Ease; RDFKGL: Flesch-Kincaid Grade Level; RDL2: L2 Readability.}
    \label{tab:data-features}
\end{table}

\section{Rater Training Procedures}\label{appendix-training}

To familiarize the raters with the assessment procedures, we arranged an online training session via a video conferencing software. Two authors of this study introduced the source texts and corresponding reference interpretations, and clarified certain key terms within the analytic rating scales (e.g. ``filled pauses'', ``long silence'', and ``excessive repairs''). Raters were actively encouraged to seek clarification on any aspect of the rating task, so as to ensure a shared understanding of the assessment criteria. To enhance rating consistency, pre-scored, representative interpretations from each band were played and analyzed collectively. This served to illustrate the typical features associated with different performance levels. Subsequently, the raters independently completed trial ratings of five additional interpretations. After that, they engaged in a collaborative discussion, comparing their scores and providing justifications for their rating decisions. The formal rating was also conducted remotely, with each rater receiving secure online access to all necessary materials, including the source texts, reference translations, and the anonymized interpretations. To ensure ample time for thorough evaluation, raters were given two weeks to complete their assessments.

\begin{table}[h]
    \centering
    \begin{tabular}{cccccc}
\toprule
Dimension&  &  Infit MnSq  &  Outfit MnSq  &  Rater reliability  &  Person Separation reliability  \\
\midrule
 &  Rater 1  & 1.02 & 1.01 &  &  \\
InfoCom &  Rater 2  & 0.84 & 0.79 & 0.97 & 0.83 \\
 &  Rater 3  & 0.78 & 0.65 &     &     \\
\rowcolor{gray!10} &  Rater 1  & 1.15 & 1.08 &  &  \\
\rowcolor{gray!10}FluDel &  Rater 2  & 1.04 & 1.01 & 0.98 & 0.81 \\
\rowcolor{gray!10} &  Rater 3  & 0.89 & 0.75 &     &     \\
 &  Rater 1  & 1.07 & 0.99 &  &  \\
TLQual &  Rater 2  & 0.90 & 0.93 & 0.96 & 0.76 \\
 &  Rater 3  & 0.82 & 1.04 &     &     \\
\bottomrule
    \end{tabular}
    \caption{Infit, outfit, rater reliability, and person separation reliability statistics from the MFRM analysis.}
    \label{tab:rater-reliability}
\end{table}

\clearpage
\section{Prompt for Chinese Grammatical Error Diagnosis}\label{appendix-prompt}

\newcommand{\promptbox}[2]{%
\begin{tcolorbox}[enhanced jigsaw,breakable,pad at break*=1mm,
  colback=black!5!white,colframe=black!75!black,title=#1]
#2
\end{tcolorbox}
}
\promptbox{Prompt for Chinese grammatical error diagnosis}{
**Instruction**

You are a Chinese grammar expert. Your task is to diagnose and correct grammatical errors in Chinese sentences or longer texts. Follow the steps and guidelines below meticulously:\\

1. Error Detection and Analysis Order

Analyze the input text for potential errors in the following priority order:

  - Redundancy (R): Repeated words or characters that unnecessarily clutter the sentence.

  - Missing Words (M): Omitted words or particles that make the sentence incomplete or ambiguous.

  - Word Selection (S): Inappropriate or inaccurate word choices that should be replaced by more context-appropriate terms.

  - Word Order (W): Incorrect arrangement of words or phrases that distorts the intended meaning.\\

2. Error Description and Correction

For each detected error:

  - Describe the nature of the error.
  
  - Propose a correction that clarifies the meaning while preserving the original intent.
  
  - Assign a confidence score (0–1) representing your certainty in the correction. (Scores closer to 1 indicate high confidence.)\\

3. Re-examination for Low Confidence

If an error receives a confidence score below 0.7, re-examine it by asking:

  - ``Does this correction improve the sentence without introducing ambiguity?''
  
  - ``Is the error type correctly classified?''

Revise the correction if necessary before finalizing your output.\\

4. Handling Special Cases

The following special cases should be addressed:

  - Filled Pauses: Words such as ``\begin{CJK*}{UTF8}{gkai}呃\end{CJK*}'', ``\begin{CJK*}{UTF8}{gkai}额\end{CJK*}'', and ``\begin{CJK*}{UTF8}{gkai}嗯\end{CJK*}'' (and similar utterance markers) are considered fillers and should be ignored during error analysis. Do not report these as grammatical errors.
  
  - Repeated Phrases, False Starts, and Self-Corrections: Only analyze the final output of the sentence. Ignore any extraneous parts resulting from repetition or self-correction.\\

5. Output Formatting

For every detected error, output an entry using the following format:

[sentence\_id, start\_index, end\_index, error\_type, corrected\_text, confidence]

  - sentence\_id: A unique identifier for the sentence (or text segment) under analysis.
  
  - start\_index and end\_index: The character positions (based on the sentence's index) where the error occurs.
  
  - error\_type: One of the following codes: R (Redundancy), M (Missing Words), S (Word Selection), or W (Word Order).
  
  - corrected\_text: The proposed correction.
  
  - confidence: A numerical value between 0 and 1 that represents your certainty.\\

6. Multiple Errors

Note that a sentence or text passage may contain more than one error. In such cases, output each error as a separate entry.\\

7. Examples for Illustration

- Example 1: Simple Redundancy Correction

  \quad- Input: \begin{CJK*}{UTF8}{gkai}我昨天去学校学校了。\end{CJK*}
  
  \quad- Expected Output: [1, 6, 7, R, \begin{CJK*}{UTF8}{gkai}学校\end{CJK*}, 0.95]

  \quad- Reasoning: The particle ``\begin{CJK*}{UTF8}{gkai}了\end{CJK*}'' is repeated unnecessarily (positions 6–7). The extra ``\begin{CJK*}{UTF8}{gkai}了\end{CJK*}'' should be removed. High confidence is given due to the unambiguous redundancy.

- Example 2: Word Order Correction

  \quad- Input: \begin{CJK*}{UTF8}{gkai}他跑得快比我还。\end{CJK*}
  
  \quad- Expected Output: [2, 4, 6, W, \begin{CJK*}{UTF8}{gkai}比我还快\end{CJK*}, 0.85]

  \quad- Reasoning: The phrase ``\begin{CJK*}{UTF8}{gkai}跑得快比我还\end{CJK*}'' is mis-ordered. Reordering to ``\begin{CJK*}{UTF8}{gkai}比我还快\end{CJK*}'' aligns with natural Chinese word order.

- Example 3: Word selection improvement

  \quad- Input: \begin{CJK*}{UTF8}{gkai}不受监管的移民活动会造成移民进入许多危险的路线, 也会遭到人口贩卖者的残忍魔爪。\end{CJK*}
  
  \quad- Expected Output:
  
   \quad\quad Entry 1: [3, 10, 11, S, ``\begin{CJK*}{UTF8}{gkai}让\end{CJK*}'', 0.95]
   
   \quad\quad Entry 2: [3, 25, 28, S, ``\begin{CJK*}{UTF8}{gkai}移民会落入\end{CJK*}'', 0.90]
  
  \quad- Reasoning: Entry 1: ``\begin{CJK*}{UTF8}{gkai}造成\end{CJK*}'' is not a suitable verb.
Entry 2: ``\begin{CJK*}{UTF8}{gkai}遭到\end{CJK*}'' is not a natural collocation with ``\begin{CJK*}{UTF8}{gkai}魔爪\end{CJK*}''. The verb ``\begin{CJK*}{UTF8}{gkai}落入\end{CJK*}'' better conveys that immigrants ``fall into'' the clutches (\begin{CJK*}{UTF8}{gkai}魔爪\end{CJK*}) of human traffickers. Additionally, the extra adverb ``\begin{CJK*}{UTF8}{gkai}也\end{CJK*}'' is unnecessary.

- Example 4: Handling Special Cases

   \quad- Input: \begin{CJK*}{UTF8}{gkai}呃，我觉得今天的会议，嗯，没啥大问题。\end{CJK*}
   
   \quad- Expected Output: No error entries.
   
   \quad- Reasoning: ``\begin{CJK*}{UTF8}{gkai}呃\end{CJK*}'' or ``\begin{CJK*}{UTF8}{gkai}嗯\end{CJK*}'' are neglected. Only the final phrasing after self-corrections and filler pauses should be examined for genuine grammatical issues.
}

\clearpage
\section{Complete Feature Statistics}\label{appendix-features}
\begin{table}[h!]
    \centering
    \adjustbox{width=0.78\textwidth,center}{
    \begin{tabular}{lcccccccc}
\toprule
\multirow{2}{*}{Feature} &  \multicolumn{2}{c}{Mean} &  \multicolumn{2}{c}{SD} &  \multicolumn{2}{c}{Skewness} &  \multicolumn{2}{c}{Kurtosis} \\
 &  Raw  &  Aug.  &  Raw  &  Aug.  &  Raw  &  Aug.  &  Raw  &  Aug.  \\
\midrule
\rowcolor{gray!10} \multicolumn{9}{c}{InfoCom features} \\
 CometKiwi  & 0.51 & 0.51 & 0.10 & 0.06 & 0.13 & 0.22 & -0.53 & 0.82 \\
 BertScore  & 0.96 & 0.96 & 0.01 & 0.00 & -0.73 & -1.20 & -0.32 & 1.56 \\
 chrF  & 0.11 & 0.11 & 0.02 & 0.02 & 0.14 & 0.16 & -0.55 & 1.23 \\
 BLEURT-20  & 0.51 & 0.50 & 0.13 & 0.07 & 1.14 & 1.87 & 2.85 & 1.52 \\
 XCOMET  & 0.18 & 0.17 & 0.11 & 0.06 & 1.06 & 1.66 & 1.77 & 0.96 \\
\midrule
\rowcolor{gray!10} \multicolumn{9}{c}{FluDel features} \\
 NUP  & 34.05 & 34.57 & 14.95 & 15.26 & 0.78 & 0.73 & 1.24 & 2.16 \\
 MLUP  & 1.00 & 0.94 & 0.61 & 0.46 & 2.01 & 2.53 & 5.35 & 7.62 \\
 MLFP  & 0.35 & 0.35 & 0.14 & 0.08 & -0.06 & -0.12 & 0.80 & 5.01 \\
 NFP  & 15.72 & 15.40 & 8.41 & 6.03 & 0.68 & 0.51 & 0.89 & 1.29 \\
 MLR  & 16.99 & 17.00 & 2.60 & 1.58 & 0.53 & 0.58 & 1.04 & 2.35 \\
 PSC  & 197.78 & 196.12 & 55.36 & 34.18 & 0.76 & 0.84 & 0.55 & 0.97 \\
 PTR  & 0.63 & 0.59 & 0.12 & 0.09 & 0.18 & 0.19 & -0.76 & 0.80 \\
 MLS  & 0.26 & 0.26 & 0.04 & 0.02 & 0.96 & 1.17 & 3.64 & 1.72 \\
 SR  & 1.73 & 1.72 & 0.48 & 0.39 & 0.81 & 0.87 & 1.58 & 1.91 \\
 AR  & 3.87 & 3.86 & 0.53 & 0.32 & 0.03 & 0.07 & 2.08 & 1.61 \\
 NRSA  & 3.75 & 3.76 & 2.99 & 1.82 & 1.25 & 1.57 & 1.72 & 0.58 \\
 NPSA  & 0.78 & 0.80 & 1.28 & 1.16 & 2.28 & 2.27 & 5.75 & 2.33 \\
 NRLFP  & 0.18 & 0.17 & 0.54 & 0.42 & 3.51 & 5.37 & 9.03 & 7.39 \\
 NRLUP  & 1.05 & 0.99 & 1.33 & 1.21 & 1.81 & 1.98 & 4.02 & 3.26 \\
\midrule
\rowcolor{gray!10} \multicolumn{9}{c}{TLQual features} \\
 NRW  & 1.68 & 1.70 & 0.51 & 0.55 & 0.44 & 0.12 & 1.23 & 2.57 \\
 NMW  & 2.17 & 2.15 & 0.62 & 0.67 & -0.43 & -0.30 & 2.26 & 1.95 \\
 NWSE  & 4.13 & 4.16 & 1.15 & 1.48 & 1.33 & 0.68 & 2.34 & 3.39 \\
 NWOE  & 0.98 & 1.02 & 0.34 & 0.36 & 0.88 & 0.57 & 1.95 & 2.64 \\
 MLC  & 16.87 & 16.84 & 2.70 & 2.46 & 0.79 & 0.79 & 1.47 & 5.38 \\
 MLTU  & 19.57 & 20.04 & 3.46 & 3.87 & 0.98 & 1.05 & 1.34 & 2.32 \\
 NCPS  & 3.69 & 3.68 & 1.28 & 1.64 & 1.49 & 2.35 & 3.58 & 2.64 \\
 NTPS  & 3.20 & 3.27 & 1.11 & 1.55 & 1.35 & 2.17 & 2.70 & 1.44 \\
 TOTAL\_RTTR  & 5.41 & 5.45 & 0.94 & 0.81 & 0.18 & -0.10 & 1.18 & 3.61 \\
 VO\_RATIO  & 0.21 & 0.22 & 0.08 & 0.04 & 0.18 & -0.11 & 1.15 & 2.76 \\
 VO\_RTTR  & 2.55 & 2.58 & 0.62 & 0.54 & -0.22 & -0.70 & 2.01 & 2.23 \\
 SP\_RATIO  & 0.22 & 0.23 & 0.09 & 0.11 & 0.81 & 0.68 & 3.94 & 3.50 \\
 SP\_RTTR  & 2.54 & 2.52 & 0.60 & 0.52 & -0.06 & -0.40 & -0.19 & 1.24 \\
 AN\_RATIO  & 0.08 & 0.09 & 0.04 & 0.02 & 0.24 & -0.14 & 3.08 & 1.12 \\
 AN\_RTTR  & 1.48 & 1.51 & 0.65 & 0.47 & -0.64 & -1.27 & 2.29 & 1.16 \\
 AP\_RATIO  & 0.37 & 0.39 & 0.09 & 0.05 & -0.02 & -0.48 & -0.69 & 2.86 \\
 AP\_RTTR  & 3.18 & 3.19 & 0.77 & 0.62 & 0.21 & -0.06 & 3.40 & 1.09 \\
 CN\_RATIO  & 0.01 & 0.01 & 0.02 & 0.01 & 1.71 & 1.63 & -0.30 & 0.82 \\
 CN\_RTTR  & 0.40 & 0.42 & 0.58 & 0.44 & 0.98 & 0.70 & 2.39 & 1.18 \\
 PP\_RATIO  & 0.03 & 0.03 & 0.03 & 0.02 & 1.61 & 2.18 & -1.45 & 1.32 \\
 PP\_RTTR  & 0.67 & 0.71 & 0.56 & 0.39 & -0.15 & -0.69 & 4.48 & 2.71 \\
 PV\_RATIO  & 0.04 & 0.05 & 0.04 & 0.02 & 1.78 & 2.01 & 5.64 & 2.63 \\
 PV\_RTTR  & 0.89 & 0.89 & 0.57 & 0.41 & -0.41 & -1.05 & -0.79 & 1.96 \\
 PC\_RATIO  & 0.04 & 0.04 & 0.04 & 0.03 & 1.22 & 1.32 & 1.41 & 3.62 \\
 PC\_RTTR  & 0.88 & 0.91 & 0.64 & 0.71 & -0.26 & -0.82 & -1.01 & 2.78 \\
\bottomrule
    \end{tabular}
    }
    \caption{Descriptive statistics of all extracted features on raw data and augmented data.}
    \label{tab:full-statistics}
\end{table}

\clearpage
\section{Case Studies of Model Prediction Errors}\label{appendix-case}

\begin{table}[ht]
    \centering
    \adjustbox{width=1\textwidth,center}{
    \rowcolors{1}{white}{gray!10}
    \begin{tabular}{m{3cm}m{13cm}}
\toprule
Sample 47  &  From the original dataset; RF model True score: 6.34; Predicted score: 5.29  \\
Key features  &  BLEURT: 0.66; CometKiwi: 0.62; chrF: 0.07; BERTScore: 0.97; xCOMET: 0.35  \\
Key features (M±SD) for Score 6 samples  &  BLEURT (0.54±0.13); CometKiwi (0.54±0.10); chrF (0.13±0.02); BERTScore (0.96±0.01); xCOMET (0.21±0.12)  \\
Error analysis  &  The model underestimates the InfoCom score of Sample 47 by 1.05. Upon examining samples within the 5.5–6.5 score range, we observe that Sample 47 exhibits a particularly low chrF score (0.07). This value is more than one standard deviation below the mean (0.11) for this feature among samples in this range. Analysis of the corresponding student transcript reveals a tendency to reorder sentence components during interpretation, though key information in the source speech is interpreted faithfully into the target language. For instance, when interpreting an ``if...then...'' sentence, the student processes the ``then'' clause before the ``if'' clause, which results in reduced n-gram matching and consequently a lower chrF score for this sample.  \\
\bottomrule
    \end{tabular}
    }
    \caption{Cases of notable disagreement between machine and human scores for InfoCom.}
    \label{tab:case-infocom}
\end{table}

\begin{table}[h]
    \centering
    \adjustbox{width=1\textwidth,center}{
    \rowcolors{1}{white}{gray!10}
    \begin{tabular}{m{3cm}m{13cm}}
\toprule
Sample 95  &  From the original dataset; XGBoost model True score: 4.73; Predicted score: 3.48  \\
Sample features  &  NFP: 13; MLR: 20.64; MLUP: 1.18; NUP: 42; MLFP: 0.26; PSC: 185; SR: 1.53; PTR: 0.41; NRSA: 2; MLS: 0.25  \\
Features (M±SD) for Score 5 samples  &  NFP (18.16±5.66); MLR (17.13±1.11); MLUP (1.02±0.11); NUP (30.4±6.73); MLFP (0.38±0.12); PSC (195.96±13.44); SR (1.72±0.25); PTR (0.45±0.24); NRSA (4.4±3.55); MLS (0.27±0.04)  \\
Error analysis  &  For Sample 95, the model underestimates the FluDel score by 1.25 points. Analysis of this sample's features reveals notably high values for MLUP (1.18) and NUP (42), both approximately two standard deviations above their respective means. Also, the speech rate (1.53) is lower than the mean (1.72). Collectively, these feature values likely lead the model to interpret this sample as having more significant breakdowns and reduced speaking speed. However, qualitative examination of the corresponding student recording offers a contrasting perspective. While the student does exhibit longer and more frequent pauses than average, these disfluencies predominantly occur at boundaries between semantic units within sentences. For human rates, this placement of pauses does not hurt perceived fluency as much as within-phrase disfluencies, which may explain why the actual perceived score is higher than the model's prediction based on these automated features.  \\
\bottomrule
    \end{tabular}
    }
    \caption{Cases of notable disagreement between machine and human scores for FluDel.}
    \label{tab:case-FluDel}
\end{table}

\begin{table}[h]
    \centering
    \adjustbox{width=1\textwidth,center}{
    \rowcolors{1}{white}{gray!10}
    \begin{tabular}{m{3cm}m{13cm}}
\toprule
Sample 62  &  From the original dataset; XGBoost model True score: 6.22; Predicted score: 5.01  \\
Key features  &  CN\_RATIO: 0; PC\_RTTR: 0; MLS: 19.57; PP\_RTTR: 1; SP\_RTTR: 0.71; AP\_RTTR: 2; MLC: 14; NWSE: 0.26; PV\_RTTR: 0.89; MLTU: 17.11  \\
Key features (M±SD) for Score 6 samples  &  CN\_RATIO (0.01±0.01); PC\_RTTR (0.99±0.39); MLS (21.36±7.18); PP\_RTTR (0.81±0.29); SP\_RTTR (2.54±0.35); AP\_RTTR (3.23±0.44); MLC (17.08±1.30); NWSE (1.69±0.74); PV\_RTTR (0.98±0.34); MLTU (19.73±1.56)  \\
Error analysis  &  The predicted score is 1.21 points lower than that assigned by human raters. A contributing factor to this discrepancy may be the notable absence of two specific Chinese structures, CN and PC expressions, in the student's interpretation. Instead, the students frequently employ expressions characteristic of Westernized Chinese, a style influenced by Western language structures. While human raters appear to find these alternative expressions acceptable within the context of the task, the model likely penalizes the lack of the expected native Chinese forms, leading to the observed lower scores.  \\
\bottomrule
    \end{tabular}
    }
    \caption{Cases of notable disagreement between machine and human scores for TLQual.}
    \label{tab:case-TLQual}
\end{table}



\end{document}